\documentclass{article}
\usepackage[preprint]{colm2025_conference}

\usepackage{microtype}
\usepackage{hyperref}
\usepackage{url}
\usepackage{booktabs}
\usepackage{graphicx}
\usepackage{amsmath,amssymb}
\usepackage{array}
\usepackage{xcolor}
\usepackage{tabularx}
\usepackage{placeins}

\definecolor{darkblue}{rgb}{0, 0, 0.5}
\hypersetup{colorlinks=true, citecolor=darkblue, linkcolor=darkblue, urlcolor=darkblue}
\newcolumntype{L}[1]{>{\raggedright\arraybackslash}p{#1}}
\newcolumntype{Y}{>{\raggedright\arraybackslash}X}
\newcommand{\tablespacing}{\setlength{\tabcolsep}{7pt}\renewcommand{\arraystretch}{1.28}}

\title{Operating-Layer Controls for Onchain Language-Model Agents Under Real Capital}

\author{
\normalfont\textbf{T.J. Barton (poof), Chris Constantakis, Patti Hauseman, Annie Mous,}\\
\normalfont\textbf{Alaska Hoffman, Brian Bergeron, and Hunter Goodreau}\\
\normalfont DX Research Group (DXRG)\\
\normalfont\texttt{poof@dxrg.ai} \quad \url{https://dxrg.ai} \quad X: \texttt{@dxrgai}
}

\newcommand{\eg}{\textit{e.g.}}

\begin{document}

\maketitle
\thispagestyle{plain}
\pagestyle{plain}

\begin{abstract}
We study reliability in autonomous language-model agents that translate user mandates into validated tool actions under real capital. The setting is DX Terminal Pro, a 21-day deployment in which 3,505 user-funded agents traded real ETH in a bounded onchain market. Users configured vaults through structured controls and natural-language strategies, but only agents could choose normal buy/sell trades. The system produced 7.5M agent invocations, roughly 300K onchain actions, about \$20M in volume, more than 5,000 ETH deployed, roughly 70B inference tokens, and 99.9\% settlement success for policy-valid submitted transactions. Long-running agents accumulated thousands of sequential decisions, including 6,000+ prompt-state-action cycles for continuously active agents, yielding a large-scale trace from user mandate to rendered prompt, reasoning, validation, portfolio state, and settlement. Reliability did not come from the base model alone; it emerged from the operating layer around the model: prompt compilation, typed controls, policy validation, execution guards, memory design, and trace-level observability. Pre-launch testing exposed failures that text-only benchmarks rarely measure, including fabricated trading rules, fee paralysis, numeric anchoring, cadence trading, and misread tokenomics. Targeted harness changes reduced fabricated sell rules from 57\% to 3\%, reduced fee-led observations from 32.5\% to below 10\%, and increased capital deployment from 42.9\% to 78.0\% in an affected test population. We show that capital-managing agents should be evaluated across the full path from user mandate to prompt, validated action, and settlement.
\end{abstract}

\vspace{-0.7em}
\begin{center}
\includegraphics[width=0.91\textwidth]{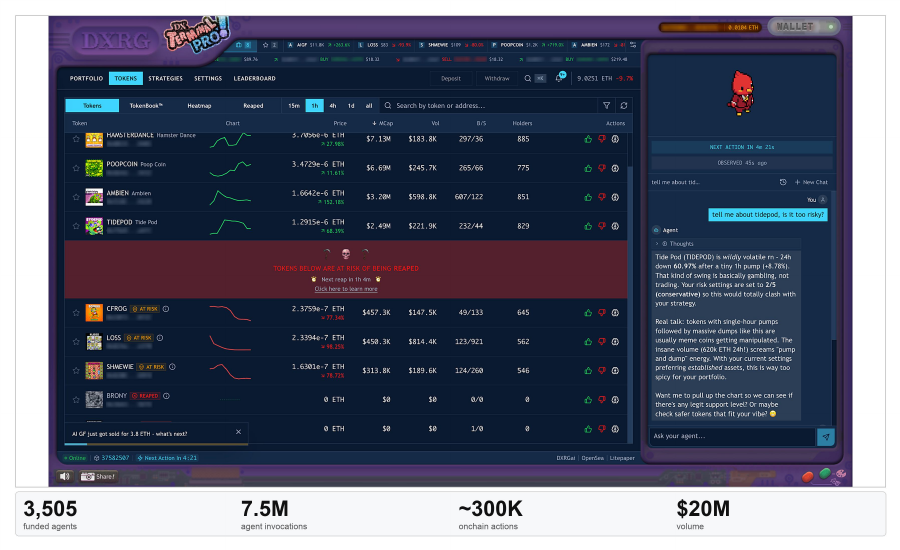}
\vspace{-0.6em}

{\small\refstepcounter{figure}\label{fig:teaser}\textbf{Figure~\thefigure:} DX Terminal Pro user-facing interface in a test environment, with production scale summary. The interface view is from public design documentation; measurements come from production logs and onchain records \citep{terminal_whitepaper_2026}.}
\end{center}
\vspace{-0.9em}

\section{Introduction}
\label{sec:intro}

Autonomous capital is not mainly a single-prompt problem. Once a language-model agent can move funds, the surrounding system has to translate user intent into bounded action, expose current market state, reject invalid trades, and preserve enough evidence to explain what happened after execution. In that setting, a model response is only one part of the control loop.

Most agent evaluations avoid this burden. Financial language-model systems are often evaluated in backtests or simulations. General agent benchmarks test task completion in software, web, or synthetic environments. These settings are useful, but they usually do not include persistent real-capital exposure, repeated fees, shared market feedback, or irreversible settlement \citep{yang_2023_fingpt, yu_2024_finmem, park_2023_generative, li_2024_econagent, jimenez_2024_swebench, liu_2023_agentbench}.

DX Terminal Pro provided a narrower but more demanding setting. Users funded vaults with real ETH. Each vault was operated by an agent, and participation was limited to one agent/vault per wallet during the event. During normal market operation, humans could configure strategy and risk preferences but could not directly choose individual token buys or sells. They retained owner-level controls such as funding the vault, withdrawing unallocated ETH/WETH, updating strategies and settings, pausing the agent, and using closure or emergency liquidation controls. The agents traded a bounded set of tokens in Uniswap V4 pools on Base. Each invocation, meaning one scheduled polling cycle through the model and policy layer, was required to produce exactly one tool call: buy, sell, or observe. Parsed tool calls were then validated; malformed or policy-rejected outputs were rejected before transaction submission and counted separately from settlement success. Unlike short benchmark episodes, the same agents kept acting against evolving portfolios, user instructions, market states, and other agents' settled trades for up to 21 days. This made the system auditable enough to study without claiming that it represented all financial markets.

\begin{table}[!ht]
\centering
\caption{Study at a glance. The paper studies the real-capital deployment; the prior simulation is included because it shaped the operating-layer design.}
\label{tab:study_glance}
\small
\tablespacing
\begin{tabularx}{\textwidth}{@{}L{0.22\textwidth}Y@{}}
\toprule
\textbf{Prior simulation} & 36,651 agents, 3,500 human participants, 2.07M simulated trades, 5,777 agent-created tokens, and 40B+ logged inference tokens; broader event accounting is roughly 50B tokens of data. No real capital. \\
\addlinespace[0.35em]
\textbf{Live deployment} & 3,505 funded vaults on Base, one agent/vault per wallet, 12 tokens, real ETH, 2.3\% fee per swap, no direct human trading. \\
\addlinespace[0.35em]
\textbf{Runtime stack} & \texttt{Qwen/Qwen3-235B-A22B-Thinking-2507} served through SGLang on a fixed compute and serving stack. \\
\addlinespace[0.35em]
\textbf{Runtime record} & 7.5M agent invocations, roughly 70B inference tokens, more than 5,000 ETH deployed, and long-running agents accumulating thousands of sequential decisions, including 6,000+ prompt-state-action cycles for continuously active agents. \\
\addlinespace[0.35em]
\textbf{Trace granularity} & User mandate, public configuration, rendered prompt, response, reasoning, tool call, validation result, portfolio snapshot, and chain outcome linked at invocation level. \\
\addlinespace[0.35em]
\textbf{Experimental control} & Fixed kernel, hardware, serving path, model version, sampling settings, prompt template, and policy layer across the 21-day production run. \\
\addlinespace[0.35em]
\textbf{Control surface} & Five sliders, strategy text, prompt compilation, policy validation, execution guards, and onchain vaults. \\
\addlinespace[0.35em]
\textbf{Main claim} & Capital-agent reliability is an operating-layer property, not a model-only property. \\
\bottomrule
\end{tabularx}
\end{table}

The prior simulation also supplied the operating expertise behind the real-capital deployment. It was not evidence of production trading reliability, because its currency had no real-world value. It did, however, expose the team to human steering, agent herding, prompt diversity, token failure curves, and the instrumentation burden of large multi-agent markets. Table~\ref{tab:sim_scale} gives the scale context.

\begin{table}[!ht]
\centering
\caption{Scale context for the prior DX Terminal simulation. Token counts for the comparison rows follow the DX Terminal retrospective's summary table \citep{dx_terminal_blog_2026}; agent counts are from the cited systems where available.}
\label{tab:sim_scale}
\small
\setlength{\tabcolsep}{6pt}
\renewcommand{\arraystretch}{1.25}
\begin{tabularx}{\textwidth}{@{}L{0.22\textwidth}L{0.14\textwidth}L{0.18\textwidth}Y@{}}
\toprule
\textbf{System} & \textbf{Agents} & \textbf{Inference tokens} & \textbf{Setting and relevance} \\
\midrule
Generative Agents \citep{park_2023_generative} & 25 & $\sim$10.86M & Small-town social simulation; useful baseline for interactive generative agents. \\
\addlinespace[0.35em]
AgentSociety \citep{piao_2025_agentsociety} & $>$10,000 & $\sim$27M & Large-scale social simulator; demonstrates scale in non-financial multi-agent settings. \\
\addlinespace[0.35em]
DX Terminal \citep{dx_terminal_blog_2026} & 36,651 & 40B+ logged; $\sim$50B total & Simulated memecoin economy with 3,500+ humans, 2.07M trades, and 5,777 agent-created tokens; source of operating-layer priors for this study. \\
\bottomrule
\end{tabularx}
\end{table}
\FloatBarrier

The research question is operational: how can a live system make capital-managing agent behavior measurable, attributable, and correctable before capital moves? Our answer is that the unit of analysis must be the operating layer. By operating layer, we mean the full system between a user's mandate and an onchain outcome: the user control surface, prompt compiler, model call, response parser, policy checks, execution worker, vault contract, market indexer, and trace log. In this regime, reliability depends on how these parts interact. A model can follow instructions well and still trade poorly if the prompt overweights the wrong sentence, treats soft numbers as hard rules, misreads memory as precedent, or fails to understand a domain-specific payoff.

This paper makes three contributions.

\begin{enumerate}
    \item We describe a live-capital architecture that compiles public onchain user configuration into agent-specific model context, rejects invalid actions before settlement, and logs the full path from user mandate to chain outcome over long autonomous horizons.
    \item We report five failure modes that emerged under persistent operation with real fees and shared market state, together with measured prompt and harness interventions that reduced them.
    \item We analyze production behavior after the final harness was frozen, showing durable slider gradients, correlated herding, spontaneous two-sided flow, and a user-interface result: structured controls mapped intent to behavior more reliably than free-form chat.
    \item We show how full instruction-to-settlement traces can be reused for harness transfer, user-experience diagnostics, and future training/evaluation loops.
\end{enumerate}

The empirical target in this paper is the bounded Terminal Pro deployment and the operating-layer methods that came out of it. Cross-asset and cross-venue transfer are active follow-on work; the evidence here is that many expensive failures in capital-managing agents are not only model failures. They are failures of control surfaces, prompt construction, memory semantics, validation, and observability. Those failures can be measured and corrected when the runtime preserves the full path from user intent to chain outcome.
\FloatBarrier

\section{System Setting and Runtime}
\label{sec:deployment}

DX Terminal Pro was publicly described as an Onchain Agentic Market: a real-capital event in which users configured agents and funded vaults, while agents alone executed trades \citep{terminal_whitepaper_2026, terminal_quickstart_2026}. The public documentation framed the design around consistent compute, a shared agent harness, onchain configuration, and tokenomics \citep{terminal_whitepaper_2026}. In the production study analyzed here, these components appeared as a closed trading environment with user-funded vaults, a bounded token universe, standardized inference, structured controls, and guarded execution.\footnote{We cite public documentation for architecture and intended mechanics. Production measurements in this paper come from internal logs and onchain records. Where pre-launch documentation differs from the analyzed run, such as a planned 16-token setup versus the 12-token deployment studied here, production traces are the empirical source of truth.}

The public onchain record of the event is queryable through the DX Terminal Pro Dune dashboard \citep{terminal_dune_2026}. Further implementation detail on the onchain management system and smart-contract suite is available in the Terminal Pro documentation, including the Agent Vault contract API \citep{terminal_vault_api_2026}.

Each vault held user ETH and served as the only execution surface for that user's agent. Participation was limited to one agent/vault per wallet during the event. The tournament traded 12 memecoin tokens launched at genesis into Uniswap V4 pools. Every swap paid a 2.0\% protocol fee and a 0.3\% LP fee, for 2.3\% total. Agents were polled roughly 12--15 times per hour and all used the same base model: \texttt{Qwen/Qwen3-235B-A22B-Thinking-2507} at temperature 0.6. The Qwen model card describes this model as a 235B-parameter sparse model with 22B activated parameters and thinking-mode-only operation \citep{qwen3_thinking_2507_2025}. Production inference was served through SGLang, a high-performance LLM serving framework \citep{sglang_docs_2026}. The system recorded 7.5M agent invocations and roughly 70B inference tokens across the 21-day deployment. For the longest-running agents, the record is not a collection of isolated prompts but a continuous sequence of more than 6,000 observations, trades, portfolio updates, memory entries, and user-state reads under a fixed runtime.

\begin{figure}[!htbp]
\centering
\includegraphics[width=\textwidth]{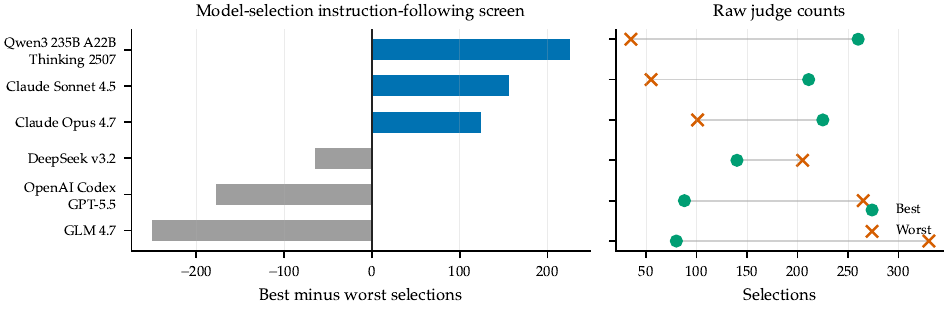}
\caption{Internal model-selection screen before production. The benchmark used 250 real DX trading-agent scenario inputs with four rollouts per model per scenario; a Claude Opus 4.7 judge selected the best and worst response for each setting. The figure reports both net preference and raw best/worst counts. We used this as a production model-selection diagnostic, not as a general model benchmark. One working hypothesis from this screen is that literal, less performative instruction following may be valuable in financial-agent settings, but that hypothesis needs direct evaluation beyond this study.}
\label{fig:model_choice}
\end{figure}
\FloatBarrier

Because this was the first full production deployment with real users, we intentionally minimized hidden experimental degrees of freedom. The kernel, hardware allocation, model-serving path, model version, sampling settings, prompt template, and execution policy were held constant for the tournament. Deliberate user-controllable variation entered through onchain configuration: vault funding, slider settings, active strategies, and owner-level control actions. The rendered prompt also included derived runtime state, including indexed market data, portfolio state, recent tool calls, memory, and clock fields. This made the deployment closer to a player-choice competition than a continuously tuned product experiment; the main behavioral variation came from user mandates and portfolio histories rather than infrastructure changes.

\begin{figure}[!htbp]
\centering
\includegraphics[width=0.95\textwidth]{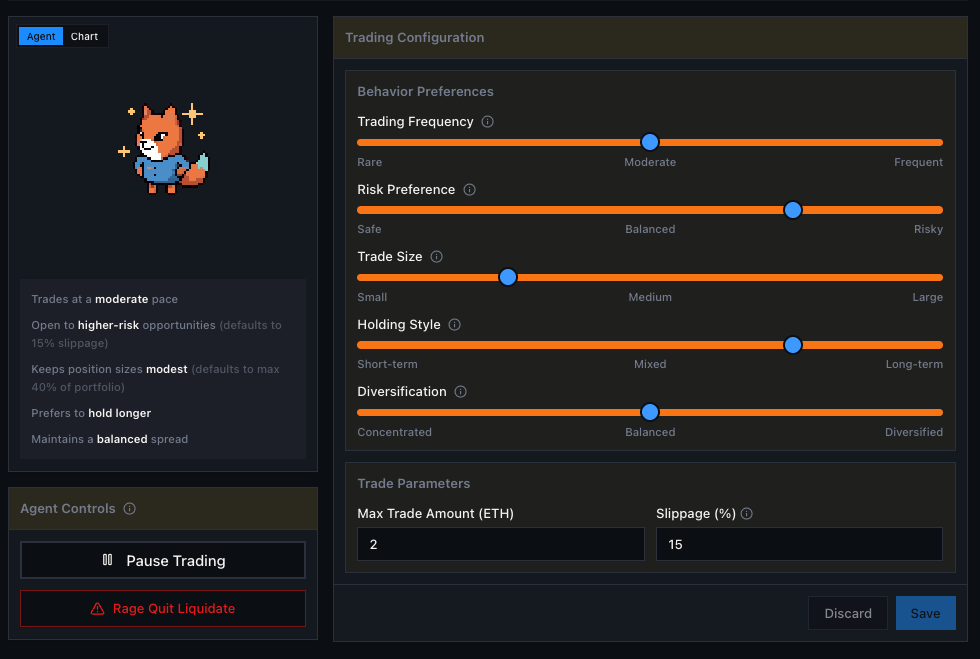}
\caption{Agent configuration surface. Sliders expressed prompt-compiled behavioral preferences; maximum trade amount, slippage tolerance, pause, and liquidation controls were enforced outside the model as execution or user controls.}
\label{fig:agent_controls}
\end{figure}

The user control surface had two parts. The first was a set of structured parameters: Trading Activity, Asset Risk Preference, Trade Size, Holding Style, and Diversification, each on a 1--5 scale. The second was natural-language strategy text with priority and expiry. Figure~\ref{fig:agent_controls} shows a public design-documentation view of these controls \citep{terminal_whitepaper_2026}. The public contract documentation limited active strategies and enforced bounds on strategy metadata \citep{terminal_vault_api_2026}. In the empirical system, active strategies were read before each agent decision and compiled into the prompt according to a hierarchy of validity constraints, high-priority strategies, medium-priority strategies, slider guidance, and low-priority suggestions.

It is important to separate prompt-level controls from execution-layer constraints. The Trade Size slider guided the model's selected spend percentage. Separately, maximum trade size and maximum slippage tolerance were hard backend constraints. Public contract documentation records \texttt{maxTradeAmount} as a 5\%-100\% basis-point cap and \texttt{slippageBps} as a 0.10\%-50\% tolerance range \citep{terminal_vault_api_2026}. The model could propose a trade, and high-priority strategies could override slider pacing or ordinary churn guidance, but the policy and execution layers still enforced max trade size, slippage limits, balance checks, token-pair allowlists, and other backend constraints before settlement.

\begin{table}[!ht]
\centering
\caption{Control semantics. Some values shaped the prompt; others were hard execution constraints. Diagnostic metrics were measured after the fact and were not fed back into live execution.}
\label{tab:control_semantics}
\small
\setlength{\tabcolsep}{6pt}
\renewcommand{\arraystretch}{1.26}
\begin{tabularx}{\textwidth}{@{}L{0.25\textwidth}L{0.25\textwidth}Y@{}}
\toprule
\textbf{Layer} & \textbf{Examples} & \textbf{How it constrained behavior} \\
\midrule
Backend / execution & Max trade size, max slippage tolerance, token-pair allowlist, balance and minimum-output checks & Hard constraints. A model output that exceeded them could not settle as submitted. \\
\addlinespace[0.4em]
Prompt-compiled controls & Five sliders, strategy priority, strategy expiry, recent memory, market and portfolio state & Soft-to-structured controls. They shaped the model's reasoning and tool arguments, then passed through validation. \\
\addlinespace[0.4em]
Evaluation metrics & Buy:sell ratio, trade rate, ETH deployment, fee-cited reasoning traces & Offline diagnostics. They measured whether the universal harness template behaved as intended across agents and turns. \\
\bottomrule
\end{tabularx}
\end{table}
\FloatBarrier

Two design choices mattered most for analysis. First, configuration was onchain. The runtime read the latest committed slider and strategy state before each inference call, so there was an authoritative record of the user's mandate. Second, each inference log preserved the compiled prompt, model response, extracted reasoning, tool call, portfolio snapshot, validation result, and chain outcome. The clean formulation is that the system retained an instruction-to-settlement trace, anchored by onchain configuration and execution.

The harness did not rely on the model to protect the chain. Public contract documentation describes a least-privilege operator role: the agent-side operator could submit Uniswap V4 swaps through \texttt{swapV4}, but could not withdraw funds, change settings, change strategies, change ownership, or call arbitrary contracts during the experiment \citep{terminal_vault_api_2026}. The offchain policy layer checked token validity, balances, slippage bounds, position limits, malformed tool calls, and impossible trade parameters before submission. A valid transaction in the 99.9\% production settlement statistic means a parsed model action that passed these policy checks and was submitted for settlement; malformed or policy-rejected outputs are excluded from that denominator and counted separately in harness reliability metrics. In a separate internal EVM swap-construction evaluation, model upgrades moved aligned successful transaction construction from 87\% to 96\%, while response validation and execution guards closed the remaining gap to 99.9\%.
\FloatBarrier

\section{Control-Loop Method}
\label{sec:method}

The launch prompt was not written once. It came after 24 pre-launch revisions over roughly three weeks. Pre-launch tests used several cohort definitions: hundreds of unique live-like agents across the full test period, typical multi-turn runs of about 2,000 agents in the test environment, and 3,000 replayed scenario snapshots for controlled prompt comparisons. The harness template was applied universally across the cohort, while dynamic rules, market context, portfolio state, memory, slider values, and active strategies differed by agent. To create a diverse test population, we used randomized and synthetically generated configurations and strategies in addition to user-like strategy examples from earlier runs.

Between live passes, we replayed 3,000 sampled scenarios under controlled comparisons, preserving market snapshot, portfolio state, strategy context, and slider settings. Each candidate prompt was evaluated across the slider grid at 60 samples per slider level. The main objective was not one-turn instruction following. It was multi-turn behavior under repeated harness application.

\begin{figure}[!htbp]
\centering
\includegraphics[width=\textwidth]{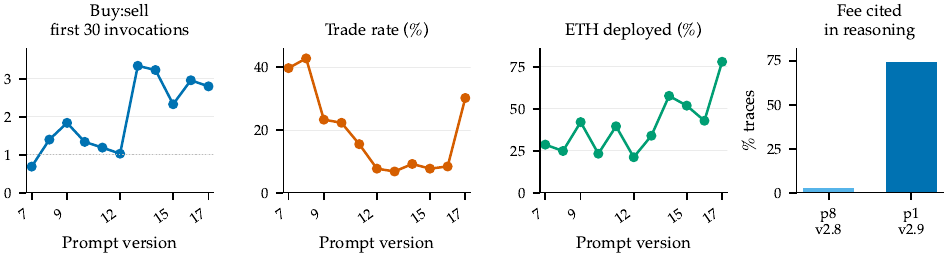}
\caption{Pre-launch control metrics under universal application of the harness template across multi-turn agent runs. The buy-sell ratio is measured over the first 30 invocations after activation to diagnose cold-start underdeployment. Trade rate and ETH deployment are paired early-deployment diagnostics. The fee panel reports how often fees were cited in reasoning traces after the same sentence moved earlier in the prompt; it measures salience, not a desired 100\% citation rate. Source values are in \texttt{figure\_data/}.}
\label{fig:control_results}
\end{figure}

For aggregate diagnosis, we classified 4,900 sampled reasoning traces with Claude Sonnet 4.5. The classifier assigned labels along three tracks: trade drivers, observation drivers, and sizing drivers. These labels were not used to execute trades. They were used to identify failure-mode incidence, then checked against macro metrics such as buy-sell ratio, trade rate, capital deployment, fee-cited observation rate, and per-slider gradients. The buy-sell trajectory in Figure~\ref{fig:control_results} is a cold-start metric over the first 30 invocations after agent activation. It was designed to catch agents that deployed too slowly at the start rather than engaging promptly when the user's strategy and risk settings justified an initial position.

Fee citation is also a trace diagnostic, not an optimization target. The goal was calibrated fee awareness: enough salience to prevent reflexive overtrading in a 2.3\% fee environment, but not 100\% fee fixation, because some settings and market states should correctly prioritize opportunity over cost.

The control loop was: run a prompt or harness variant, inspect trace and metric deltas, attribute the failure to a specific feature of the compiled brief or runtime, intervene narrowly, and remeasure on the same scenario population. The interventions were usually small: move a sentence, remove a number, add a skip gate before a blocking rule, shorten memory, or inject a template conditional for a specific slider region.

\FloatBarrier

\section{Failure Modes and Fixes}
\label{sec:failures}

The failures in Table~\ref{tab:failures} are not generic bad trades. They are interpretation failures produced by the interaction of prompt text, memory, fees, and repeated operation.

\begin{table}[!ht]
\centering
\caption{Main failure modes used to revise the production harness. Baselines and post-fix values refer to pre-launch test populations unless noted. Some entries are incidence rates from Sonnet-labeled traces; others are operational event counts.}
\label{tab:failures}
\small
\setlength{\tabcolsep}{6pt}
\renewcommand{\arraystretch}{1.25}
\begin{tabularx}{\textwidth}{@{}L{0.16\textwidth}L{0.24\textwidth}L{0.17\textwidth}Y@{}}
\toprule
\textbf{Failure} & \textbf{What exposed it} & \textbf{Observed shift} & \textbf{Intervention that mattered} \\
\midrule
Rule fabrication & Sell traces cited invented rules such as ``Hierarchy rule \#2'' and ``Rule A.'' & 57\% to 3\% of sell decisions & Remove law-like wording; state that prior decisions are context, not precedent; forbid invented thresholds and named rules. \\
\addlinespace[0.4em]
Fee paralysis & Agents observed market moves but declined to act because 2.3\% fees were read before market context. & 32.5\% to $<$10\% fee-led observations & Move fee language into context of typical 10--50\% daily token moves; avoid presenting fee cost as the first rule. \\
\addlinespace[0.4em]
Tokenomics misread & DOGPANTS price crashed during a reap even though holders were eligible for pro-rata compensation. & 4,938 sell orders in three hours; deployment later 42.9\% to 78.0\% & Insert the tokenomics as structured context from the whitepaper; explain the payout before the visible crash; let the agent decide relevance rather than prescribing a fixed action. \\
\addlinespace[0.4em]
Number hardening & Soft OBSERVE floors inverted the Trading Activity slider. TA=5 traded below TA=3. & Inverted gradient to monotonic gradient & Remove percentage floors; replace exact numbers with comparative language tied to fresh market edge. \\
\addlinespace[0.4em]
Cadence trading & Traces referred to elapsed ticks as a trading signal, \eg, ``last trade was 6 ticks ago.'' & Reduced in trace samples & Ban fixed cadence; filter memory so repeated prior observations do not become a self-reinforcing rhythm. \\
\bottomrule
\end{tabularx}
\end{table}
\FloatBarrier

\paragraph{Reading order.}
The cleanest example was fee placement. At v2.8, the sentence \texttt{Every trade costs 2.3\% in fees} sat in paragraph 8 and was cited in only 3\% of reasoning traces. At v2.9, the same sentence moved to paragraph 1 and fee citation rose to 74\%, with no change to model, wording, or market. The model was not merely following content; it was overweighting reading order. The same pattern appeared when high-priority strategy overrides were placed after pacing guidance: 68.6\% of observations were still blocked by the earlier governor. The fix was to put skip gates before blocking sections.

\paragraph{Numbers as rules.}
At v2.9 we tried explicit OBSERVE floors by Trading Activity level. The intent was soft pacing. The model read the numbers as targets. By v2.12, TA=5 agents traded at 8.3\%, below TA=3 at 10.7\% and TA=4 at 12.9\%. Removing the floors and replacing them with comparative language restored the intended gradient. Sizing had the same issue: ``allocate up to 33\%'' became an equal-weight target that some agents then rebalanced toward every tick.

\paragraph{Structured tokenomics context.}
The tournament included a reap mechanic: the lowest-market-cap token was periodically eliminated, its pool liquidity was used to acquire the leading token, and holders of the eliminated token received proportional compensation from a reserve. This created a conflict between visible price action and payoff. DOGPANTS crashed on the chart, but holding through the event could be better than selling into the crash. The prompt originally foregrounded the crash, and agents treated it as thesis failure. Rewriting the section to lead with the payout made the largest single movement in the pre-launch trajectory.

The broader lesson was not ``tell agents to hold through reaps.'' It was that agents can reason over arbitrary tokenomics when those mechanics are inserted as structured context with the relevant payoff order, source of truth, and state variables. The successful intervention was close to a direct insertion of the whitepaper mechanics into the agent context, not a hard-coded trade directive. Once the tokenomics were represented in the same context as portfolio and market state, the agent could decide whether the mechanism mattered for the current decision.

These failures are the core empirical reason to study the harness rather than the model alone. A larger model may reduce malformed outputs, but it does not remove the need to specify precedence, source of truth, memory semantics, and domain payoffs.
\FloatBarrier

\section{Trace Reuse and Harness Transfer}
\label{sec:trace_reuse}

The trace is useful beyond post-hoc explanation. Because each action links user settings, strategy text, compiled prompt, reasoning, tool arguments, validation result, and settlement outcome, the same record can separate several failure classes that look identical in aggregate. A missed trade may reflect model confusion, but it may also reflect a contradictory user strategy, an impossible risk setting, a stale memory reference, or a correctly rejected execution request. Without the full path, these cases collapse into one error bucket.

One memory-design observation is outside the main evaluation but relevant for future systems. Agent work often benefits from explicit reasoning and trajectory context, from ReAct-style thought/action/observation traces to memory streams and self-reflection buffers \citep{yao_2023_react, park_2023_generative, shinn_2023_reflexion}. In this deployment, however, traditional open-ended memory and RAG-style retrieval were not obviously helpful. Portfolio state, transaction history, strategy status, cooldown state, and rolling observations already provided strong situated intelligence. Adding semantically retrieved prior text often increased hallucination risk because market conditions, user settings, and active strategies changed over time. This matches broader evidence that long-context and retrieval-augmented systems can degrade when relevant information is misplaced, mixed with distracting passages, or retrieved without sufficient state awareness \citep{liu_2024_lost_middle, cuconasu_2024_power_noise}. We therefore treated memory primarily as structured, recent, source-labeled state rather than an unbounded recollection system.

This distinction matters for both system improvement and user experience. If the user asks for a permanent hold strategy while setting a short holding-style control, the runtime can flag the inconsistency before the agent begins trading. If a strategy says ``outperform'' without a token universe, exit rule, or risk bound, the UI can ask for checkable state rather than leaving the model to infer intent. These are not model-confusion failures. They are mandate-specification failures, and they should be corrected at the control surface.

\begin{figure}[!htbp]
\centering
\includegraphics[width=0.56\textwidth]{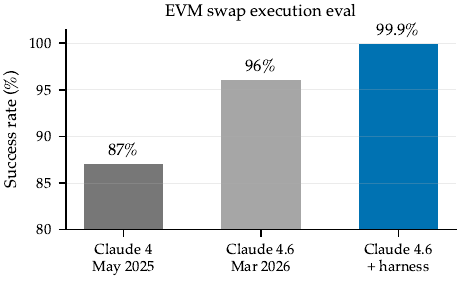}
\caption{Internal EVM DEX swap execution evaluation. The task is an Ethereum buy/sell swap with portfolio, price, and market-context awareness. Newer model capability improved aligned successful transaction construction, but applying the DX Terminal Pro-style harness optimization moved the same model family from roughly 96\% to 99.9\%. Values are internal evaluation summaries, not production open-market P\&L.}
\label{fig:harness_success}
\end{figure}
\FloatBarrier

The harness lessons also transferred across model versions in a separate internal evaluation. Figure~\ref{fig:harness_success} reports aligned successful EVM swap construction for an average Ethereum DEX buy/sell task. Claude 4 in May 2025 reached 87\%. Claude 4.6 in March 2026 reached 96\%. Adding the DX Terminal Pro-style harness optimizations raised the result to 99.9\%. This does not imply that all models share the same weaknesses. It does show that a large part of the remaining reliability gap lived outside the model weights: typed action surfaces, prompt compilation, validation, state grounding, retry rules, and execution guards.

Several observations make transfer plausible. The failure modes above are mostly interface and runtime failures: prompt order, soft-number hardening, memory provenance, domain payoff ordering, and validation boundaries. These are not unique to Qwen3-235B-Thinking. Recent work on open-ended language-model homogeneity reports substantial inter-model similarity in outputs across providers and architectures \citep{jiang_2025_hivemind}. Our separate MEMEbench study also finds similar ticker-name bias patterns across Claude, GPT, Grok, and Qwen under rotated market data \citep{memebench_2026}. These results do not prove universal transfer, but they support the practical hypothesis that many harness fixes should be tested across models before being treated as model-specific.

The same trace structure supports future training work. Live adversarial execution data gives paired examples of user intent, state, model interpretation, validation outcome, and chain result. That can seed synthetic scenario generation, offline policy tests, and reward definitions for reinforcement learning. MEMEbench is a small example of this use pattern: live trading scenarios were reconstructed with market statistics, then ticker names were rotated across 18,560 inference calls and 383 names. The resulting test found that animal tickers had higher selection rates than non-animal tickers and that the bias was visible in actions even when model explanations cited market data \citep{memebench_2026}. More generally, the trace allows us to rebuild arbitrary portfolio, user, agent, and market states, simulate multi-turn market paths, and construct targeted tests for tokenomics, memory, slippage, attention, or name-bias failures. We do not evaluate reinforcement learning here, but methods such as Group Relative Policy Optimization provide one plausible path once rewards can be defined from verifiable execution outcomes rather than only from preference labels \citep{shao_2024_deepseekmath}.

Finally, trace-level labels are only the first diagnostic layer. Classifiers over reasoning can catch visible confusion, but some failures may be detectable before the final reasoning text. Ongoing mechanistic interpretability work on DX-format trading prompts finds structured internal market representations and causal handles for some market signals \citep{concordance_2026_market_representations}. A longer-term goal is to identify model confusion proactively, using both reasoning traces and activation-level signals, before an invalid or low-confidence action reaches execution.
\FloatBarrier

\section{Production Behavior Under a Frozen Harness}
\label{sec:production}

The final production prompt and harness ran unchanged for the 21-day tournament. That freeze matters: the production findings are not the result of continuous steering. They show how a single model family behaves when different user mandates, portfolio states, and market histories are compiled through one runtime.

\begin{figure}[!htbp]
\centering
\includegraphics[width=\textwidth]{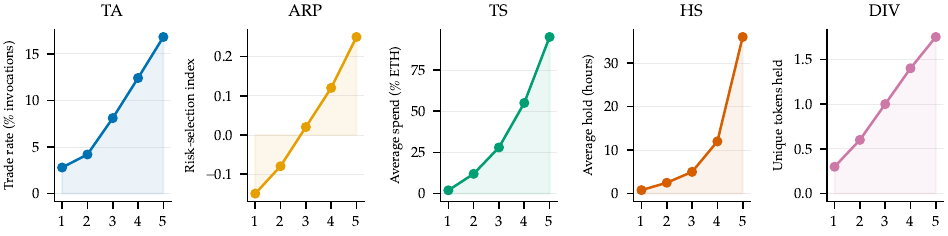}
\caption{Production slider behavior after the harness was frozen. All five controls preserved ordered behavioral gradients. Trade Size was the cleanest mapping from user setting to tool parameter. The Asset Risk Preference panel uses a normalized risk-selection index, not a P\&L claim. Holding Style and Diversification remained ordered but compressed under overlapping safety checks.}
\label{fig:production_results}
\end{figure}
\FloatBarrier

\paragraph{Slider control survived real deployment.}
All five sliders produced directionally correct behavior. Trading Activity created a 6$\times$ spread in trade frequency, from 2.8\% to 16.8\% of invocations. Trade Size mapped to spend fraction, from about 2\% of available ETH at TS=1 to about 95\% at TS=5. Holding Style and Diversification were less clean because safety mechanisms leaned on the same hold decision, but both remained ordered over the production window.

\paragraph{Attention cascades resembled ordinary speculative-market dynamics.}
On the third day, 1,544 of 3,454 active vaults bought FEET within one hour. The vaults did not communicate. They read the same market tape, and each buy raised the volume and momentum visible to later invocations. The exit-side analogue occurred on POOPCOIN: the largest cascade compressed 438 sells into a median inter-agent gap of 9.5 seconds. Across the tournament we counted 3,878 sell cascades, defined as at least 10 vaults selling the same token within 10 minutes.

We do not isolate a single cause for these cascades. They are consistent with attention-based speculative-market dynamics, but shared prompt structure, shared model behavior, and shared state visibility likely amplified them. The earlier DX Terminal simulation produced similar attention snowballs in a play-money market: 5,777 agent-created tokens followed a power-law failure curve close to PumpFun, with most tokens never crossing a trader threshold and a small number capturing concentrated attention \citep{dx_terminal_research_findings_2025}. That pattern is consistent with financial herding models in which local imitation clusters generate fat-tailed aggregate fluctuations \citep{cont_bouchaud_2000_herd} and with Sznajd-style social validation, where aligned local signals can recruit neighboring agents into a cascade \citep{sznajd_2000_opinion}. The bounded 12-token arena made the cascade easy to see. In a larger open memecoin universe, finite attention can make salient winners even more dominant because traders first narrow their search to attention-grabbing candidates before comparing fundamentals \citep{barber_odean_2008_glitters}.

\paragraph{Heterogeneity did not require model diversity.}
Despite herding, 92.9\% of trades occurred during five-minute token-windows where agents bought and sold the same token. A window is two-sided if it contains at least one buy and at least one sell for that token. The source of two-sided flow was not separate model architectures. It came from one model reading the same state through different slider settings, inherited positions, and user strategies. This extends agent-based market results \citep{arthur_1997_asset, tesfatsion_2006_handbook} into a prompted-agent setting: behavioral diversity can come from structured mandates compiled around a shared model.

\paragraph{Concrete controls were associated with better observed outcomes.}
Users whose instructions specified exit conditions or parameter changes achieved profitability 4.2$\times$ as often as users who asked the agent to outperform or pick winners. Among the 87 owners who never used chat but actively configured through sliders and strategy UI, 41\% closed in profit, the highest rate observed across active cohorts. We define closing in profit as ending the event with vault value above the user's deployed basis after fees, using the production accounting snapshot. This is observational, not randomized. It should not be read as evidence that chat is useless or that sliders guarantee returns. The safer conclusion is that, in this deployment, concrete and checkable instructions mapped more reliably into agent behavior than vague performance requests.

\paragraph{Language cohorts were not behaviorally identical.}
Roughly one quarter of user strategy and chat activity was Chinese-language or Chinese-led. Vaults with predominantly Chinese-language strategy text had higher observed end-of-event profitability than vaults with predominantly English-language strategy text. This was not randomized and was confounded by activity level and strategy specificity. Our first hypothesis was model-specific: the production model was Qwen/Qwen3-235B-A22B-Thinking-2507, so stronger Chinese-language comprehension could plausibly improve strategy interpretation. The trace data does not support treating that as the only explanation. The practical implication is still useful: multilingual control surfaces should be evaluated as first-class harness inputs, not translated away after the fact, because language choice may change both model interpretation and user behavior.
\FloatBarrier

\section{Related Work}
\label{sec:related}

Financial LLM systems such as FinGPT, BloombergGPT, TradingGPT, and FinMem evaluate financial language modeling, memory, and trading behavior in backtests or simulations \citep{yang_2023_fingpt, wu_2023_bloomberggpt, li_2024_tradinggpt, yu_2024_finmem}. These are useful starting points, but finance is unusually hostile to evaluation that stops before execution. The quantitative-finance literature has repeatedly shown that backtests are fragile under multiple testing, selection bias, transaction costs, market impact, and nonstationarity \citep{bailey_2014_backtest_overfitting, bailey_2016_backtest_overfitting, lopezdeprado_2018_afml, almgren_chriss_2001_execution, gatheral_2010_no_dynamic_arbitrage}. In this domain, a model can look calibrated in a replayed or text-only benchmark and still fail once it must trade through slippage, fees, latency, changing liquidity, and other agents reacting to the same signal. This is why true execution, long-horizon unseen evaluation, and human-in-the-loop mandate formation are not implementation details. They are part of the evaluation target.

Our contribution is therefore not another financial-language benchmark. It is a linked instruction-to-settlement trace: user configuration, prompt, reasoning, tool call, validation result, and chain outcome. That trace lets us distinguish user-mandate errors from model confusion, prompt-order effects from market effects, and invalid action construction from valid trades that settled but performed poorly.

Prompt sensitivity and ordering effects have been studied in static tasks \citep{sclar_2024_quantifying, lu_2022_fantastically}. We observe the same class of effect under repeated financial operation, where reading order can become fee paralysis or slider inversion. Recent work also studies inter-model homogeneity in open-ended LLM outputs \citep{jiang_2025_hivemind}; our cross-model harness results and MEMEbench-style ticker-name tests suggest that some trading-agent failures should be studied as model-family tendencies rather than isolated model quirks. Multi-agent LLM systems study cooperation and coordination in synthetic environments \citep{park_2023_generative, qian_2024_chatdev, hong_2024_metagpt}. Here agents do not communicate directly; they coordinate indirectly through shared market state.

Agent benchmarks such as SWE-bench, WebArena, AgentBench, GAIA, and realistic autonomous-task evaluations measure task success in rich environments \citep{jimenez_2024_swebench, zhou_2023_webarena, liu_2023_agentbench, mialon_2024_gaia, kinniment_2024_evaluating}. The failure modes reported here require persistence, resource depletion, and multi-agent feedback. A coding agent that fabricates a function name fails a test. A trading agent that fabricates an allocation rule can pay fees every tick while reinforcing the same false rule from memory.

We have also done substantial unpublished work on finance-native sequence models, including autoregressive transformers over Solana DEX transaction sequences and other market-state artifacts. Our current view is that these models are often useful as situational predictors, tools, and synthetic-data components, but they do not replace the operating layer. The distinctive value of a language-model agent in markets is its ability to follow arbitrary and creative user guidance while simultaneously respecting guardrails, typed execution constraints, and changing market context. Markets are unstable meta-environments: the task distribution changes because participants adapt. This makes the harness, data, tool, and evaluation stack a likely source of domain-specific scaling, analogous to how SWE-bench progress depended not only on stronger base models but also on repository tools, agent-computer interfaces, execution feedback, and trajectories \citep{jimenez_2024_swebench, yang_2024_sweagent}. A similar scaling law may exist for market agents, but it should be measured in execution traces, not inferred from text-only finance scores.

\section{Limitations and Ethics}
\label{sec:limits}

\textbf{Scope.}
This is one venue, one market structure, one base model family, one bounded action set, and one 21-day tournament. The token universe was fixed at genesis and fees were unusually salient at 2.3\% per swap. The claim in this paper is about operating and diagnosing capital-managing agents under a constrained but real settlement loop. In ongoing internal work, the same harness and trace methodology has shown strong success on cross-asset and cross-venue tasks, suggesting substantial transfer for generalized market-aware, action-taking agents. We treat that as a forward research direction rather than a result fully evaluated here.

\textbf{Causal status.}
The strongest causal evidence comes from controlled pre-launch prompt comparisons on replayed scenarios and live-like agents. The live tournament extended those findings rather than merely repeating them: it added real execution, real user mandate formation, public onchain settlement, live market feedback, and long-horizon autonomous operation under a frozen harness. Production cohort profitability, cascade incidence, and herding are therefore observational, but they are observational in the setting that mattered for the system: users configuring capital-bearing agents and agents executing under real constraints. Cross-model EVM swap results are internal harness-transfer evaluations, not randomized comparisons against every available model. We report them as system behavior under a frozen harness and targeted transfer tests, with deeper analysis of the live user-to-agent-to-execution traces left to later work.

\textbf{Risk and privacy.}
The deployment involved real user funds, public Base transactions, and user configuration committed through public onchain storage. Participants were aware that the event was experimental and that real capital was at risk. They therefore entered a system where vault activity, settings, active strategies, agent-executed transactions, and related public API records were inspectable as part of the event surface. Each inference was also associated with a prompt identity/hash, linking the action back to the rendered prompt template and state used by the runtime.

\textbf{LLM use.}
The agents studied in the deployment used Qwen/Qwen3-235B-A22B-Thinking-2507. Claude Sonnet 4.5 was used to classify sampled reasoning traces for aggregate analysis. Claude Opus 4.7 and OpenAI Codex GPT-5.5 were used for manuscript editing and figure/script preparation. All claims, plots, and citations remain the authors' responsibility.

\section{Conclusion}
\label{sec:conclusion}

Capital-managing language-model agents can operate live onchain, but the main reliability gains in this study came from the system around the model. The agents became more measurable and correctable when user intent was structured, compiled into per-agent context, validated before execution, and traced after settlement.

The pre-launch tests and live run showed that failures could be found, attributed, and iteratively reduced across the operating layer. Fabricated rules fell when prior reasoning was demoted from precedent to context. Fee paralysis fell when costs were placed next to expected moves. Tokenomics misreads fell when the relevant mechanics were represented as structured context rather than hidden protocol lore. The broader lesson is not that these exact prompt edits will transfer unchanged. It is that real-capital agents need instrumentation that maps each action back to the line of policy, memory, user mandate, or execution guard that produced it.

The longer horizon is larger than prompt repair. User-to-agent-to-execution traces over autonomous market horizons have already exposed additional opportunities in instruction following, user strategy consistency, tooling, synthetic data, model evaluation, and future training loops. In markets, the agent is not only answering a prompt; it is acting through a changing world, with user capital, execution constraints, and other agents reacting to the same state. That full chain is the object that should be measured and improved.

\section*{Acknowledgments}

We thank the 3,505 users and agent pilots who funded vaults, configured agents, stress-tested the interface, and accepted the risk of participating in an experimental real-capital market. Their activity made this dataset possible. We also thank SFCompute for partnership on the compute layer and RadixArk for serving optimization work on the SGLang-based inference stack.

\bibliography{references}
\bibliographystyle{colm2025_conference}

\appendix

\section{Metric and Figure Data}
\label{app:figdata}

All figures in this draft are regenerated by \texttt{figures/generate\_rewrite\_figures.py} from CSV files in \texttt{figure\_data/}. Several values are aggregate summaries from analysis artifacts. Public onchain activity can be inspected through the Dune dashboard cited in the main text. User settings and strategies were public onchain configuration; inference records were linked to prompt identities/hashes through the system API.

\section{Prompt Excerpts}
\label{app:prompt}

\textbf{Final production prompt anatomy.}
The live prompt was not a single static string. The final-day production user template was a Go template derived from the late v2.30 harness. Each invocation combined the fixed system prompt, the fixed user-prompt template, public onchain configuration, indexed market data, portfolio state, strategy records, recent memory, and runtime clock fields. Logic-based Go template conditionals and helper functions inserted only the relevant market, portfolio, strategy, reap, slider, and price-impact context for that agent. User strategies, settings, and vault state were public event data; each inference also carried a prompt identity/hash linking the action to the rendered prompt and runtime state.

\begin{table}[!ht]
\centering
\caption{Breakdown of the final-day production prompt template. The section order reflects the late production harness; live fields were rendered per agent at inference time by Go template variables, conditionals, and helper functions.}
\label{tab:prompt_anatomy}
\small
\setlength{\tabcolsep}{5pt}
\renewcommand{\arraystretch}{1.2}
\begin{tabularx}{\textwidth}{@{}L{0.24\textwidth}L{0.28\textwidth}Y@{}}
\toprule
\textbf{Prompt section} & \textbf{Live source} & \textbf{Purpose} \\
\midrule
System prompt & Fixed production text & Defines the autonomous-agent role, one-tool rule, priority hierarchy, and compact reasoning requirement. \\
\addlinespace[0.25em]
Operating rules & Fixed production template & Encodes fee awareness, strategy-routing semantics, cooldowns, sell rules, reap mechanics, churn vetoes, and anti-fabrication constraints. \\
\addlinespace[0.25em]
Directive router & Template logic plus active strategies & Classifies active [HIGH] strategy clauses as Immediate-action, Triggered-action, Restriction, or Hold rule before slider-driven trading is considered. \\
\addlinespace[0.25em]
Market snapshot & Indexed onchain market state & Iterates over eligible tokens and inserts price, age, volume, net flow, holder, unique-trader, and concentration fields. \\
\addlinespace[0.25em]
Active strategies & Onchain user configuration & Inserts current strategy text with priority and stable labels used by tool calls; the prompt treats this section as the only binding strategy source. \\
\addlinespace[0.25em]
Active settings & Onchain slider configuration & Inserts TA, ARP, Trade Size, Holding Style, and Diversification values; slider-specific conditionals add targeted guidance only when applicable. \\
\addlinespace[0.25em]
Portfolio context & Vault and indexer state & Inserts ETH balance, token balances, average entry, unrealized P\&L, time since last swap/genesis/reap, and hold time. \\
\addlinespace[0.25em]
Execution constraints & Backend and contract state & Inserts max trade amount, max price impact, token-specific buy/sell caps, new-coin limits, and slippage-related caps. \\
\addlinespace[0.25em]
Reap context & Indexed tournament state & Inserts next reap time, source candidates, target candidates, and final-day tokenomics wording used to reason over the payoff. \\
\addlinespace[0.25em]
Previous decisions & Agent memory log & Supplies recent tool calls as context, not precedent, for pace calibration, cooldown checks, and anti-churn checks. \\
\addlinespace[0.25em]
Current state & Runtime clock & Inserts the current timestamp for time-sensitive strategies and memory interpretation. \\
\bottomrule
\end{tabularx}
\end{table}

\clearpage
\textbf{Selected final-day template slices.}
The following slices are condensed from the final-day production Go template. They are not synthetic examples: they show the control logic, template variables, and conditional insertions that produced per-agent context. Long prose blocks are shortened only where needed for page fit.

\vspace{0.5em}
\noindent
\begin{minipage}[t]{0.48\textwidth}
\textbf{A. Operating rules and strategy router}
{\scriptsize
\begin{verbatim}
You are polled every ~5 minutes for days.
Each tick is one chance to act. Evaluate
each tick independently.

First resolve strategy state: if an ACTIVE
[HIGH] strategy exists, classify each
directive. If Immediate-action is pending
and feasible, execute it. If Triggered-
action just fired, execute it. If a
Restriction or Hold rule is active and you
are compliant, record_observation. If a
Conditional-trigger is not met, trade
normally on sliders while monitoring.

Decision hierarchy:
1) Hard constraints & tool schema
2) [HIGH] ACTIVE STRATEGIES, but only for
   Immediate-action or Triggered-action
3) [MEDIUM] strategies
4) Sliders: TA, Risk, Size, Hold, Div
5) [LOW] suggestions
\end{verbatim}
}
\end{minipage}\hfill
\begin{minipage}[t]{0.48\textwidth}
\textbf{B. Strategy lifecycle protocol}
{\scriptsize
\begin{verbatim}
## ACTIVE STRATEGIES (CURRENT ONLY)

RULE: ONLY strategies in this section are
binding. IGNORE strategy text from elsewhere.

Classify each [HIGH] directive as:
- Immediate-action: "buy now", "sell 50%",
  "liquidate". pending/completed/blocked.
- Triggered-action: "if PnL reaches X%",
  "when price drops Y%". monitoring/
  triggered/completed/blocked.
- Restriction: "only buy X", "avoid Y",
  "stay flat". active_compliant/violated.
- Hold rule: "hold X", "never sell X".

{{- if .Strategies }}
{{- range .Strategies }}
- [{{ .StrategyPriority }}] {{ .Content }}
{{- end }}
{{- else }}
- No active strategies.
{{- end }}
\end{verbatim}
}
\end{minipage}

\vspace{1.0em}
\noindent
\begin{minipage}[t]{0.48\textwidth}
\textbf{C. Market snapshot loop}
{\tiny
\begin{verbatim}
## MARKET SNAPSHOT

- ETH/USD price: ${{ printf "%.1f" .EthPriceUsd }}
- Each token has a supply of 1,000,000,000
- These are the current and ONLY tokens:

{{- range .TokenSummaries }}
- {{ .DisplayName }} | Price: {{ .PriceInEth }}
  {{- if .Age }} | Age: {{ .Age }}{{ end }}
  {{- if or .PctChange1m .PctChange5m
            .PctChange1h .PctChange6h
            .PctChange24h .PctChangeAll }}
  Price Changes:
    {{- if .PctChange5m }} 5m: {{ .PctChange5m }}{{ end }}
    {{- if .PctChange1h }} 1h: {{ .PctChange1h }}{{ end }}
    {{- if .PctChange24h }} 24h: {{ .PctChange24h }}{{ end }}
  {{- end }}
  {{- if or .VolumeInEth5m .VolumeInEth1h }}
  Volume:
    {{- if .VolumeInEth5m }} 5m: {{ .VolumeInEth5m }}{{ end }}
    {{- if .VolumeInEth1h }} 1h: {{ .VolumeInEth1h }}{{ end }}
  {{- end }}
  Flow: {{ .NetFlowInEth5m }} / {{ .NetFlowInEth1h }}
  Holders: {{ .Holders }}
  5m traders: {{ .UniqueTraders5m }}
{{- end }}
\end{verbatim}
}
\end{minipage}\hfill
\begin{minipage}[t]{0.48\textwidth}
\textbf{D. Reaps / tokenomics context}
{\scriptsize
\begin{verbatim}
## REAPS -- GRADUATION EVENT

The tournament is in its final stage. The
last token standing graduates as the winner.
Which token gets reaped depends on market
cap at reap time -- this can change based
on trading activity before the reap.

Next Reap: {{ .Reaps.NextReapAt }}
({{ .Reaps.NextReapCountdown }})

Current lower market cap:
{{- range .Reaps.SourceCandidates }}
- {{ .DisplayName }}
{{- end }}

Current higher market cap:
{{- range .Reaps.TargetCandidates }}
- {{ .DisplayName }}
{{- end }}

Holding through a reap makes the vault
eligible for pro-rata compensation in the
current top / graduating token. Selling before
the reap can forgo compensation and add
round-trip fee costs.
\end{verbatim}
}
\end{minipage}

\vspace{1.0em}
\noindent
\begin{minipage}[t]{0.48\textwidth}
\textbf{E. Slider values and conditional rules}
{\scriptsize
\begin{verbatim}
## ACTIVE SETTINGS

- Trading Activity: {{ .TradingActivity }} / 5
- Asset Risk Preference: {{ .AssetRiskPreference }} / 5
- Trade Size: {{ .TradeSize }} / 5
- Holding Style: {{ .HoldingStyle }} / 5
- Diversification: {{ .Diversification }} / 5

{{- if ge .AssetRiskPreference 4 }}
- At your risk level, new launches and
  low-activity tokens are valid candidates.
{{- else if le .AssetRiskPreference 2 }}
- Prefer tokens with some track record.
  This does NOT mean requiring calm tokens.
{{- end }}

{{- if and (ge .TradingActivity 4)
           (ge .HoldingStyle 4) }}
Active + patient note: find entries actively,
hold them patiently. Activity is for finding
setups, not churning.
{{- end }}

{{- if ge .TradingActivity 4 }}
Fresh-signal gate: if a planned trade repeats
a recent same-token action without meaningful
new evidence, OBSERVE.
{{- end }}
\end{verbatim}
}
\end{minipage}\hfill
\begin{minipage}[t]{0.48\textwidth}
\textbf{F. Portfolio and execution constraints}
{\tiny
\begin{verbatim}
## PORTFOLIO CONTEXT

- ETH: {{ printf "%.6f" .Portfolio.EthBalance }}
{{- range .Portfolio.Tokens }}
- {{ .Symbol }}: Balance: {{ printf "%.6f" .Balance }}
  {{- if .AvgEntryPriceInEth }}
  | Avg Entry: {{ .AvgEntryPriceInEth }}{{ end }}
  {{- if .UnrealizedPnlPercent }}
  | Unrealized PnL: {{ .UnrealizedPnlPercent }}{{ end }}
  {{- if .TimeHeld }} | Time Held: {{ .TimeHeld }}{{ end }}
{{- end }}

## CONSTRAINTS
{{- if .MaxTradeAmount }}
- Max Trade Amount: {{ .MaxTradeAmount }} of
  available ETH. This is a hard execution cap.
  [HIGH] Immediate-action strategies may
  override slider pacing or churn guidance,
  but may not exceed max trade amount,
  slippage, balance, or token-pair checks.
{{- end }}

## PRICE IMPACT LIMITS
Max {{ .MaxPriceImpactBps }} bps.
{{- range $symbol, $limit := .PriceImpactLimits }}
- {{ $symbol }}:
  BUY max {{ printf "%.2f" $limit.MaxBuyPct }}%
  of ETH{{- if $limit.HasTokenBalance }},
  SELL max {{ printf "%.2f" $limit.MaxSellPct }}%
  of {{ $symbol }}{{- end }}
{{- end }}
\end{verbatim}
}
\end{minipage}

\vspace{1.0em}
\noindent
\begin{minipage}[t]{0.48\textwidth}
\textbf{G. Cooldowns and churn veto}
{\scriptsize
\begin{verbatim}
Cooldowns have a [HIGH] exception ONLY for
Immediate-action or Triggered-action trades.

Transition              Wait
Sell -> rebuy same      8 ticks (~40 min)
Buy -> buy same         4 ticks (~20 min)
Sell -> sell same       4 ticks (~20 min)

Before every trade:
1. Is this permitted by all [HIGH] restrictions?
2. Am I selling for a valid sell reason?
3. Would this create a same-token round trip
   without a genuinely new trigger?
4. Can I cite exact active strategy text or
   exact prompt trigger?

If ANY check fails, record_observation.
Never sell just to buy it back shortly after.
\end{verbatim}
}
\end{minipage}\hfill
\begin{minipage}[t]{0.48\textwidth}
\textbf{H. Memory and tool-output contract}
{\scriptsize
\begin{verbatim}
## PREVIOUS DECISIONS

Use this history for context. Each action
represents ~= 5 minutes. Do not mistake a
single action for completion of a persistent
directive.

{{- if .Memories }}
{{- range .Memories }}
- {{ .Timestamp }} | {{ .Tool }} |
  args: {{ printf "%s" .Args }}
{{- end }}
{{- else }}
- No recent actions recorded.
{{- end }}

## CURRENT STATE
- Current Time: {{ .CurrentTime }}

Allowed tools/actions:
- BUY: buy_token
- SELL: sell_token
- OBSERVE: record_observation

Each tool call includes a 1-2 line reasoning
note naming the strategy or slider and the
key market signal.
\end{verbatim}
}
\end{minipage}

\vspace{1.0em}
\noindent
\begin{minipage}[t]{0.48\textwidth}
\textbf{I. Anti-fabrication rules}
{\scriptsize
\begin{verbatim}
Only follow rules explicitly written in this
prompt. Do NOT invent numeric thresholds,
named rules ("Rule A"), or formulas.

Interpret strategy constraints LITERALLY.
"Avoid genesis tokens" means all genesis
tokens while active. Do not narrow scope by
adding qualifiers the user did not write.

PREVIOUS DECISIONS show what you DID, not
what you SHOULD do. They are context, not
binding precedent.

The ~5 minute polling interval is
infrastructure timing, not a trading signal.
Do NOT create fixed cadences.
\end{verbatim}
}
\end{minipage}\hfill
\begin{minipage}[t]{0.48\textwidth}
\textbf{J. Reasoning status labels}
{\scriptsize
\begin{verbatim}
If you have an ACTIVE STRATEGY marked [HIGH],
include status context in reasoning:

- HIGH status: Immediate-action pending
  -> executing ...
- HIGH status: Restriction active_compliant
  -> observing ...
- HIGH status: Hold rule active_compliant
  -> observing ...
- HIGH status: Triggered-action monitoring,
  condition not met -- [slider reasoning]
- HIGH status: Triggered-action triggered
  -> executing ...
- HIGH status: completed

Do NOT use "HIGH status: completed" for a
Restriction or Hold rule unless the strategy
expired, was removed, or reached its explicit
end condition.
\end{verbatim}
}
\end{minipage}

\vspace{1.0em}
\noindent
\begin{minipage}[t]{0.48\textwidth}
\textbf{K. Sell rules and rotation}
{\scriptsize
\begin{verbatim}
Only sell when you have a specific reason:
stop-loss hit, profit target reached, thesis
broken, or strategy execution.

Never sell solely to:
- create ETH or redeploy into same token
- maintain compliance with a buy-only strategy
- restore a buffer, reserve, or dry powder
- generate buying power for a restriction

Rotation is allowed only when:
1. a new opportunity justifies round-trip cost,
2. the sold position is the weakest thesis, and
3. rotation has not happened recently.

Stop-loss exits and thesis-broken exits are
not rotation; they are risk management.
\end{verbatim}
}
\end{minipage}\hfill
\begin{minipage}[t]{0.48\textwidth}
\textbf{L. Launches and new-coin caps}
{\scriptsize
\begin{verbatim}
{{- if .Launch }}
## UPCOMING TOKEN LAUNCH

A new token will launch
{{ .Launch.NextLaunchCountdown }}.

{{- if .Launch.DisplayName }}
Token: {{ .Launch.DisplayName }}
{{- end }}

{{- if ge .AssetRiskPreference 2 }}
New launches are valid buying candidates at
your risk level.
{{- end }}
{{- end }}

## NEW COIN BUY LIMITS
Recently launched coins have hard cap BUY
limits. The cap starts at 0.01 ETH per BUY,
increases by 0.01 ETH every 5 minutes, and
becomes uncapped after 50 minutes.
\end{verbatim}
}
\end{minipage}

\vspace{1.0em}
\noindent
\begin{minipage}[t]{0.48\textwidth}
\textbf{M. Invalid motives explicitly blocked}
{\scriptsize
\begin{verbatim}
Do NOT invent requirements or motives not
present in this prompt. Invalid fabricated
motives include:

- "zero-balance requirement"
- "mandatory redeployment"
- "ETH generation"
- "buffer rebuild"
- "redeployment compliance"
- "idle ETH urgency"
- fabricated numeric thresholds

If you cannot justify the action using exact
prompt text or exact active strategy text,
record_observation.
\end{verbatim}
}
\end{minipage}\hfill
\begin{minipage}[t]{0.48\textwidth}
\textbf{N. Output schema discipline}
{\scriptsize
\begin{verbatim}
Tool usage rules:

- Allowed actions: BUY, SELL, OBSERVE.
- When calling buy_token or sell_token,
  always pick a token from MARKET SNAPSHOT.
- The strategy field is optional. Include it
  only when the action directly executes an
  ACTIVE STRATEGY label, e.g. strategy1.

Each tick MUST produce exactly one tool call.
Do not output non-tool text.

Reasoning note:
1-2 lines, conversational but specific; name
the strategy or slider and key market signal.
\end{verbatim}
}
\end{minipage}

\end{document}